\PassOptionsToPackage{svgnames}{xcolor}
\documentclass{article} 
\usepackage{iclr2025_conference,times}

\usepackage{amsmath,amsfonts,bm}

\def\eqref#1{equation~\ref{#1}}

\def\1{\bm{1}}

\DeclareMathAlphabet{\mathsfit}{\encodingdefault}{\sfdefault}{m}{sl}
\SetMathAlphabet{\mathsfit}{bold}{\encodingdefault}{\sfdefault}{bx}{n}

\usepackage{hyperref}
\usepackage{url}
\usepackage{booktabs} 
\usepackage{graphicx} 
\usepackage{colortbl}
\usepackage{multirow}
\usepackage{wrapfig}
\usepackage{caption}
\usepackage{subcaption}

\definecolor{lgray}{rgb}{0.9,0.9,0.9}
\title{MADGEN - Mass-Spec attends to De Novo Molecular generation
}

\author{Yinkai Wang, Xiaohui Chen, Liping Liu, Soha Hassoun\thanks{corresponding author}\\
Department of Computer Science\\
Tufts University\\
\texttt{\{yinkai.wang, xiaohui.chen, liping.liu, soha.hassoun\}@tufts.edu} \\
}

\iclrfinalcopy 
\begin{document}

\maketitle

\begin{abstract}
The annotation (assigning structural chemical identities) of MS/MS spectra remains a significant challenge due to the enormous molecular diversity in biological samples and the limited scope of reference databases.  Currently, the vast majority of spectral measurements remain in the ``dark chemical space'' without structural annotations.  To improve annotation, we propose MADGEN (\underline{M}ass-spec \underline{A}ttends to \underline{D}e Novo Molecular \underline{GEN}eration), a scaffold-based method for de novo molecular structure generation guided by mass spectrometry data. MADGEN operates in two stages: scaffold retrieval and spectra-conditioned molecular generation starting with the scaffold. In the first stage, given an MS/MS spectrum, we formulate scaffold retrieval as a ranking problem and employ contrastive learning to align mass spectra with candidate molecular scaffolds. In the second stage, starting from the retrieved scaffold, we employ the MS/MS spectrum to guide an attention-based generative model to generate the final molecule. Our approach constrains the molecular generation search space, reducing its complexity and improving generation accuracy. We evaluate MADGEN on three datasets (NIST23, CANOPUS, and MassSpecGym) and  evaluate MADGEN's performance with a predictive scaffold retriever and with an oracle retriever. We demonstrate the effectiveness of using  attention to integrate spectral information throughout the generation process to achieve strong results with the oracle retriever. Our code is available at  \hyperlink{https://github.com/HassounLab/MADGEN}{https://github.com/HassounLab/MADGEN}


\end{abstract}

\section{Introduction}
Metabolomics, the measurement and identification of small molecules in biological samples, plays a critical role in numerous fields, including drug discovery, biomarker discovery, and environmental science. By analyzing the molecular composition of complex biological samples, metabolomics provides insights into cellular processes, metabolic pathways, and the effects of environmental changes on biological systems. Tandem mass spectrometry (MS/MS) has emerged as a powerful, widely used analytical technique that can separate and fragment molecules within a biological sample, thus producing rich spectra that can be further analyzed to annotate the measurements within the sample \citep{kind2018identification}.  

Despite the utility of  metabolomics, assigning a chemical structural identity to a measured spectrum remains a significant challenge. 
Currently, most MS/MS spectra cannot be linked to known molecular structures due to the vast chemical diversity in biological samples and the limited scope of reference databases. Spectral databases that catalogue molecules and their measured spectra, e.g., MoNA~\citep{MoNA} and GNPS~\citep{wang2016sharing},  are used for identifying a close match to the measured spectra. However, such databases remain relatively small. Molecular databases such as PubChem~\citep{kim2016pubchem} and KEGG~\citep{kanehisa2021kegg} are often utilized to provide \textit{candidate} molecular structures when using computational methods such as SIRIUS~\citep{duhrkop2019sirius}, MLP or GNN-based approaches ~\citep{wei2019rapid, zhu2020using} to predict the molecular structure that most likely produced the measured spectrum. Despite the success of these tools and the increased size of such databases, the ``dark chemical space'' of unknown molecules remains large, and hinders the interpretation of metabolomics data. De novo molecular structure generation from mass spectra is a promising approach to overcome the limitations of database-dependent methods. 
Further, de novo generation is crucial for discovering previously unknown  molecules that play key roles in health, disease, and environmental processes.  

Our insight in addressing this challenge herein is the use of scaffolds to simplify the structure generation process. A scaffold, or core structure, is used widely in medicinal chemistry to represent core structures of bioactive compounds \citep{hu2016computational}. Such scaffolds can be modified with the addition of functional groups to enhance medicinal properties.  By focusing  on scaffold-based molecular generation in the context of annotation, we can reduce the complexity of structure generation and constrain the search space, making it more manageable and improving  accuracy. Once a scaffold is predicted for a measured spectrum, it can guide the addition of  structural elements (atoms and bonds) to the scaffold to generate the target molecule.



 We propose a scaffold-based approach to de novo molecular structure generation guided by mass spectrometry data, with a focus on evaluating performance both when the scaffold is known and when it is predicted. Our contributions are as follows:

\begin{itemize}
    \item We introduce a two-stage framework that first predicts a  scaffold from the MS/MS spectrum, from which we then generate  the target molecular structure. Given the challenges in accurately predicting the scaffold, we report performance under two settings: using the correct scaffold and using the predicted scaffold. This comparison highlights the potential and limitations of scaffold prediction in de novo molecular generation.

    \item Our method leverages fragmentation patterns in MS/MS spectra to guide scaffold prediction. While scaffold prediction is not always accurate, integrating even partially correct scaffolds  reduces the complexity of de novo generation and constraints the search space to more plausible molecular structures.


    \item The scaffold-based design also improves interpretability, as even predicted scaffolds serve as structural anchors for understanding the generated molecules. This interpretability is crucial for analyzing potential biological functions and chemical properties in practical applications.

    \item Our approach has broad applicability in metabolomics, drug discovery, and environmental analysis, where the discovery of novel metabolites, bioactive molecules, and uncharacterized compounds is essential.
\end{itemize}

\section{Related Work}

\paragraph{De novo structure generation guided by mass spectra.}
De novo molecular  generation offers a promising alternative to database-dependent methods by directly (without the use of candidate molecules from databases) predicting or generating molecular structures from mass spectrometry data. MSNovelist \citep{stravs2022msnovelist} relies on 
CSI:FingerID \citep{duhrkop2015searching} to predict molecular fingerprints from the query mass spectrum, and then uses a LSTM model to reconstruct molecules. Spec2Mol \citep{litsa2023end} employs a convolutional neural network to map MS/MS spectra to a latent space, generating molecular structures as SMILES strings.
MassGenie \citep{shrivastava2021massgenie} uses a transformer-based model trained on real and synthetic spectra to generalize to unseen compounds, leveraging transformers' strength in handling sequential data. MS2Mol \citep{butler2023ms2mol} extends these approaches with a transformer-based encoder-decoder, incorporating byte-pair encoding and precursor mass, to improve  accuracy. There were no consistent datasets that were used to evaluate these models. For example, MSNovelist is evaluated on 3,863 MS/MS spectra from the GNPS library  \citep{wang2016sharing}, while Spec2Mol is evaluated on the NIST2020 dataset. Further, not all these tools are available in the public domain. Recently, The MassSpecGym dataset \citep{bushuiev2024mass} was developed as a benchmark dataset to standardize the evaluation on de
novo generation, retrieval, and spectra simulation tasks. We utilize this dataset, and two others, to report the performance of MADGEN. We also compare our results with the best reported results so far on the MassSpecGym dataset. 

\vspace{-0.8em}

\paragraph{Generative frameworks for molecular generation.}
Generative models have become essential in molecular generation due to their ability to approximate complex distributions in the chemical space. These models, such as VAEs, GANs, and Diffusion models, treat molecules as graphs, enabling them to capture the relational properties between atoms and bonds \citep{zhu2022survey}. Structure-constrained molecular design is a key strategy in modifying an existing candidate structure with the goal of attaining improved molecular properties. A common approach is constraining molecular generation to contain a specific scaffold or a molecular fragment, e.g., \citet{podda2020deep}, \citet{li2019deepscaffold}, \citet{green2021deepfrag}. These models often allow for an arbitrary scaffold as an initial structure that captures a desired property.  Unlike these models, MADGEN employs Murcko scaffolds \citep{bemis1996properties}, a standard scaffold used across many chemical and biological studies due to its ability to represent the core backbone of molecules.  As there are currently no methods to predict this scaffold for a query spectra, the first step of MADGEN predicts the scaffold from a list of candidate molecules. 
Importantly, generative models have shown value in exploring the uncharacterized chemical spaces \citep{holdijk2022path, chen2023efficient, duan2024react}. For example,  RetroBridge  \citep{igashov2023retrobridge} models the dependencies between the spaces of substrate and product molecules in the context of chemical reactions as a stochastic process between two distributions.  RetorBridge uses a Markov bridge process to approximate  dependencies between these intractable distributions. RetroBridge is adapted for MADGEN's second step, where we aim to model the joint distribution of scaffolds and target molecules, and starting with the Murcko scaffold, we utilize the mass spectrum to guide the generation process towards the target molecule.

\section{Methods}\label{sec3}
\begin{figure}[t]
    \centering
    \begin{tabular}{c}
         \includegraphics[width=\linewidth]{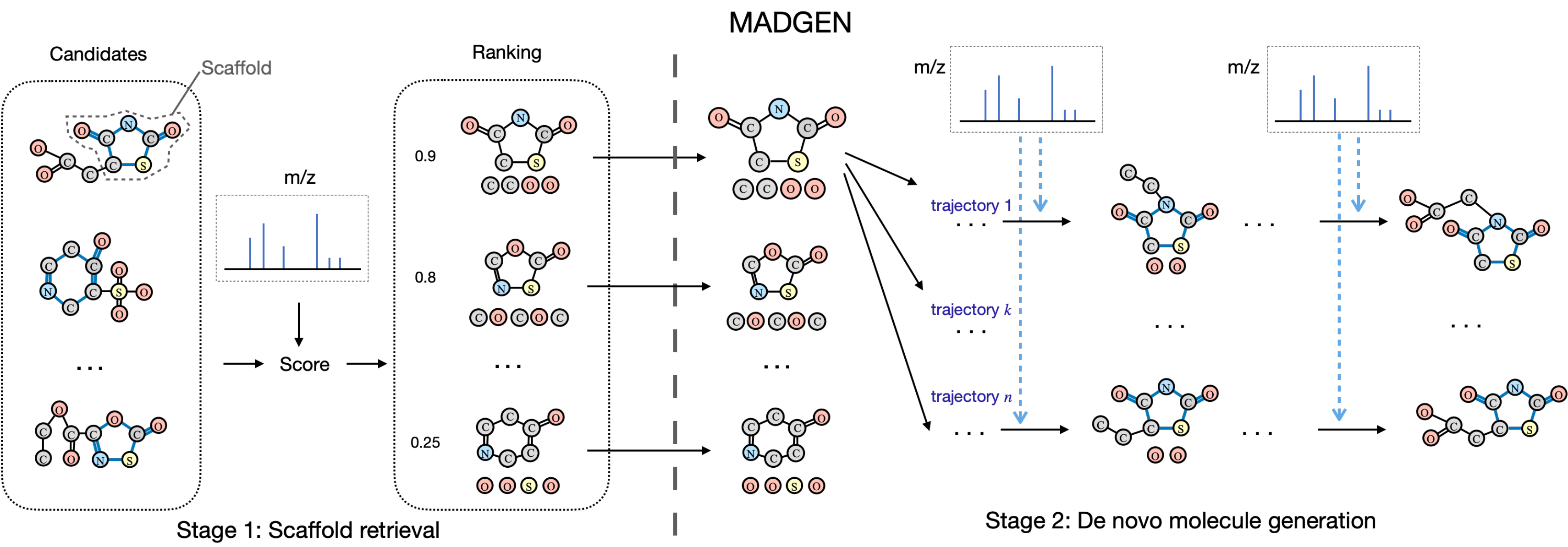}\\(a)\\ 
         \includegraphics[width=1\textwidth]{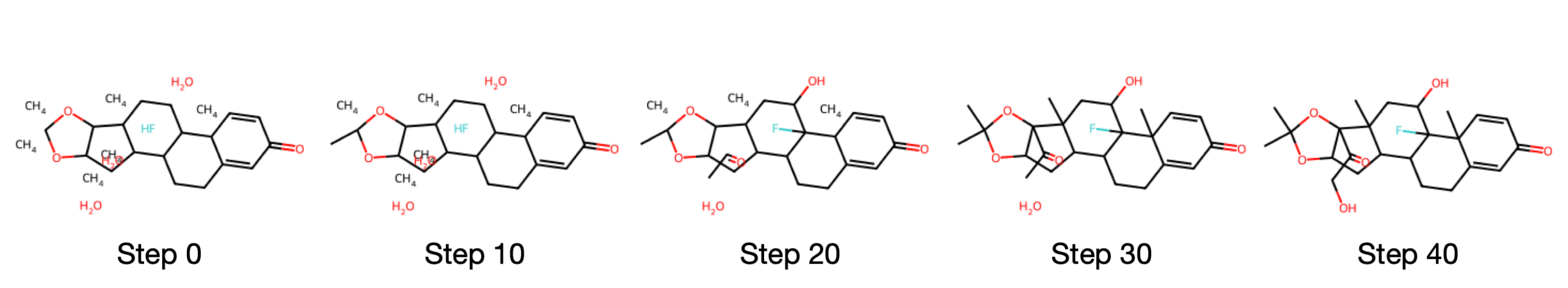}
         \\\includegraphics[width=1\textwidth]{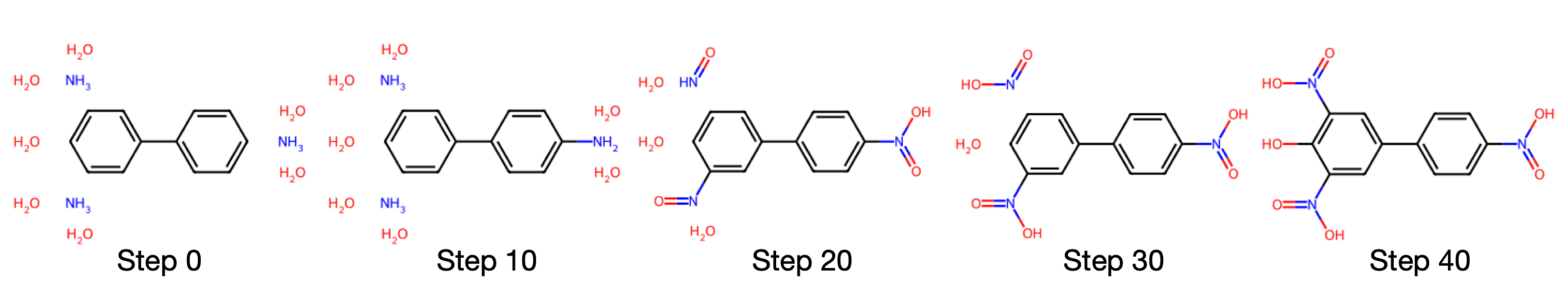}\\(b)
    \end{tabular}
       \caption{MADGEN overview and example. (a) 
       The overview of MADGEN. The mass spectra are used to rank scaffold candidates through contrastive learning. The top-ranked scaffold, with blue edges fixed, serves as a foundation for de novo molecule generation, guided by the spectra at each generation step. (b) Examples of molecular generation process over time steps for Kenalog from CANOPUS dataset (upper) and 2,6-Dinitro-4-(4-nitrophenyl)phenol from NIST23 dataset (lower). 
    The scaffolds remain fixed, while additional edges are introduced in each step to connect free atoms to scaffolds. The complete molecules are shown in step 40. 
       }\vspace{-1em}
    \label{fig:madgen_overview}
\end{figure}

Direct generation of molecules from mass spectra is a hard problem. In this work, we propose to divide the problem into two simpler sub-problems (see Figure~\ref{fig:madgen_overview}): we first retrieve the molecular scaffold from the mass spectrum and then generate the target molecule conditioned on both the mass spectrum and the scaffold. We conjecture that the scaffold prediction problem is easier than predicting the target molecule because the scaffold usually has a  simpler structure than the target molecule. Consequently, the molecule generation task becomes  easier when the scaffold is known. 

\subsection{Scaffold Retrieval}\label{sec:subsec3}
The goal of scaffold retrieval is to identify the scaffold of the target molecule. 
Denote an MS/MS spectrum and its chemical formulate as ${X}=({X}^\mathrm{ms}, {X}^\mathrm{cf})$. Scaffold retrieval takes ${X}$ as input and retrieves the core scaffold that represents the fundamental backbone of the molecule, including its ring systems and central framework. With a correct scaffold served as the starting point for further molecular generative process, the complexity of the search space is significantly reduced. 

However, predicting the scaffold from spectral data is a challenging problem due to the non-linear relationship between fragmentation patterns and the scaffold substructures. In this work, we explore two scaffold retrieval strategies - {predictive retrieval} and {oracle retrieval}. 
\vspace{-0.5em}
\paragraph{Predictive retrieval.}
We formulate the scaffold retrieval as a ranking problem. Given a set of scaffold candidates $\mathbb{S}$, the goal is to use a neural network to score each candidate $S\in\mathbb{S}$ given $X$ such that scaffold with highest score $S^*$ maximally resembles the correct scaffold $S^\mathrm{gt}$. We rank on a candidate set from which the target molecules have been removed, introducing the possibility that the true scaffold may not be present in the set.

A straightforward approach is to directly train a binary classifier that tells whether the given pair $(X, S)$ is matched or not. However, to fully leverage the relationship between the spectrum and scaffold modalities, we adopt a contrastive learning framework similar to CLIP~\citep{radford2021learning}. In this framework, the spectrum $X$ is treated as one modality, while the scaffold $S$ is treated as the other. Contrastive learning aligns the embeddings of these two modalities in a shared latent space, enabling the model to learn a meaningful representation of their relationships.

This paradigm has been widely employed in multimodal information retrieval frameworks~\citep{luo2021clip4clip,bain2022clip, lei2021less,fang2021clip2video,ma2022x,hendriksen2022extending}, where embedding similarity is used to determine the most likely paired item based on a query. Similarly, in our framework, we align the embeddings of mass spectra and scaffolds to facilitate scaffold retrieval. Specifically, we employ contrastive learning techniques inspired by JESTR~\citep{kalia2024jestr}, which was designed to align the embeddings of mass spectra with their corresponding molecules.



%

To achieve this alignment, we introduce two separate encoders to project the mass spectra and scaffold graphs into a shared latent space. Specifically, the mass spectra \( X \) are projected using a multi-layer perceptron (MLP) encoder \( f_X \), which maps the spectral data into a \( d \)-dimensional latent space. We employ a graph neural network (GNN) \( f_S \) to encode the global representation of the scaffold graph \( S \), also into a \( d \)-dimensional space.

The key insight of this approach is to ensure that the embeddings of the matched spectrum and scaffold are close to each other in the latent space.  
Both encoders, \( f_X \) and \( f_S \), are jointly trained using a contrastive learning objective. This objective ensures that the embeddings of matched spectrum-scaffold pairs are close in the joint embedding space, while mismatched pairs are pushed apart. Specifically, for each spectrum \( X \) and scaffold \( S \), we compute a similarity score, \( h(z^{n}_{\text{spec}}, z^{m}_{\text{mol}}) \), defined as:
\begin{align}
h(z^{n}_{\text{spec}}, z^{m}_{\text{mol}}) = \exp\left( \frac{ z^{n}_{\text{spec}} \cdot z^{m}_{\text{mol}} }{\| z^{n}_{\text{spec}} \| \| z^{m}_{\text{mol}} \| \tau} \right),
\end{align}

where \( z^{n}_{\text{spec}} \) and \( z^{m}_{\text{mol}} \) are the embeddings of the spectrum and molecular scaffold, respectively, and \( \tau \) is a temperature hyperparameter that controls the importance of non-matching pairs.

The contrastive loss \( \mathcal{L}_{\text{contrastive}} \) is computed over a batch of size \( k \) as:
\begin{align}
\mathcal{L}_{\text{contrastive}} = \frac{1}{k} \sum_{n=1}^{k} \left[ - \mathbb{E} \left[ \log \frac{ h(z^{n}_{\text{spec}}, z^{n}_{\text{mol}}) }{ \sum_{m=1}^{k} h(z^{n}_{\text{spec}}, z^{m}_{\text{mol}}) } \right] \right].
\end{align}
Here the two embeddings in the numerator are from the same molecule $n$ while the two in the denominator are from two different molecules. This loss encourages the model to assign high values to matching spectrum-scaffold pairs and lower values to non-matching pairs, effectively aligning the embeddings in the latent space.

After training, we access the score of each scaffold candidate via cosine similarity between the mass spectra embedding \( f_X(X) \) and the scaffold embedding \( f_S(S) \). To improve scaffold ranking accuracy, we introduce a frequency-based aggregation approach. For each data point, we first retrieve the top-$k$ ranked candidate scaffolds. The frequency of each scaffold appearing in these top candidates is then computed, and the most frequently occurring scaffold for each formula is selected as the predicted scaffold. This method refines scaffold selection by leveraging consensus among top-ranked candidates, leading to improved scaffold prediction accuracy (SPA).



\paragraph{Oracle retrieval.} We maintain a lookup table as an oracle which always yields the correct scaffold given the MS/MS spectrum and the chemical formula. We construct the lookup table by extracting the scaffold from the molecular graph representation using RDKit.This lookup table serves as an idealized oracle, simulating perfect scaffold retrieval. It allows us to focus on assessing the second stage of molecular generation: the task of adding side chains and functional groups to the scaffold-independently from any potential errors that could occur in scaffold retrieval.

\subsection{Scaffold-conditioned De Novo Molecule Generation with Spectra Guidance}\label{sec:stage2}

\subsubsection{Notations and Problem Formulation} We represent a molecule $G$ as a graph \( G = (\mathcal{V}, \mathcal{E}) \). Its scaffold \( S = (\mathcal{V}^S, \mathcal{E}^S) \) is a subgraph of \( G \). Since the atom set \( \mathcal{V} \) can be directly inferred from the chemical formula, the task of molecular generation involves determining the appropriate edge set \( \mathcal{E} \setminus \mathcal{E}^S \) that connects the scaffold to the remaining isolated atoms \( \mathcal{V} \setminus \mathcal{V}^S \). While there are combinatorially many valid edge sets that could complete the molecule from the scaffold, we utilize spectral data \( X \) to guide the edge generation process and ensure the structure aligns with the observed spectra.

\subsubsection{Scaffold-conditioned Generation via Markov Bridge}
We frame the molecule prediction task as generating graphs given a scaffold. Specifically, starting from a scaffold $S$, we are interested in modeling the distribution $p(G|S) = p(\mathcal{E}|\mathcal{E}^S, \mathcal{V}^G)$ with the following Markov decomposition:
\begin{align}
    p\big(\mathcal{E}\big|\mathcal{E}^S,\mathcal{V}^G\big)= \sum_{\mathcal{E}_0:\mathcal{E}_{T-1}}\prod_{t=0}^{T-1}p\big(\mathcal{E}_{t+1}\big|\mathcal{E}_{t},\mathcal{E}^{S},\mathcal{V}^G\big),\label{eq:model-denoise}
\end{align}
where $\mathcal{E}_0=\emptyset$ can be considered the case where no bonds are formed from isolated atoms to others, and $\mathcal{E}_T=\mathcal{E}$. The sequence of random variables $\mathcal{E}_{0:T}$ can be viewed as progressively connecting atoms to form the final molecules.

Let $e_t$ be an arbitrary edge entry in $\mathcal{E}_t$, $e_t$ can be represented as a D-dimensional one-hot vector, with 0 class being non-edge and 1 to D-1 classes being the bond types. 
Following~\citet{d3pm}, we formulate the transition probabilities $p(e_{t+1}|e_{t},e_T)$ conditioned on the endpoint $e_T$:
\begin{align}
p\big(e_{t+1}\big|e_{t},e_T\big) = \mathrm{Categorical}\big(e_{t+1}; \mathbf{Q}_{t}(e_T)e_{t}\big),
\end{align}
where $\mathbf{Q}_t(e_T)\in\mathbb{R}^{D\times D}$ is an absorbing transition matrix conditioned on the endpoint data $e_T$~\citep{igashov2023retrobridge}. 

With the defined model, we now approximate it with a parameterized distribution:
\begin{align}
p_\theta\big(e_{t+1}\big|e_{t},\mathcal{E}^{S},\mathcal{V}^G\big) = \mathrm{Categorical}\big(e_{t+1}; \mathbf{Q}_{t}(\hat{e}_T)e_{t}\big),~\text{where}~\hat{e}_T=\mathrm{nn}_\theta(\mathcal{E}_t, \mathcal{E}_S, \mathcal{V}^G)
\end{align}
is the endpoint prediction via a neural network $\mathrm{nn}_\theta(\cdot)$. Given a pair $(S,G)$ from the dataset, we train $\mathrm{nn}_\theta(\cdot)$ by optimizing the evidence lower bound (ELBO)
\begin{align}
\mathcal{L}_\theta(S,G):=-T\mathbb{E}_{\mathcal{U}(t;0,T-1)}\mathbb{E}_{p(e_t|e_0,e_T)}\bigg[\mathrm{KL}\Big(p\big(e_{t+1}\big|e_{t},e_T\big)\big\|p_\theta\big(e_{t+1}\big|e_{t},\mathcal{E}^{S},\mathcal{V}^G\big)\Big)\bigg].
\end{align}
Here, \( p(e_t|e_0, e_T) \) represents the probability of transitioning to an arbitrary timestep \( t \) from \( T \), which can be expressed in a closed form. The detailed derivation of the ELBO and the transition distributions is provided in Appendix \ref{app:derivation}.

To obtain pairs \((S, G)\) for training, we first randomly sample a graph \( G \) from the data distribution \( p(G) \). The scaffold \( S \) of \( G \) is computed using RDKit. This results in the joint data distribution \( p(S,G) = p(G)p(S|G) \), where \( p(S|G) \) is a Dirac delta distribution that assigns all its probability mass to the scaffold \( S \) derived from \( G \).

\begin{figure}[tbp]
    \centering
    \includegraphics[width=\linewidth]{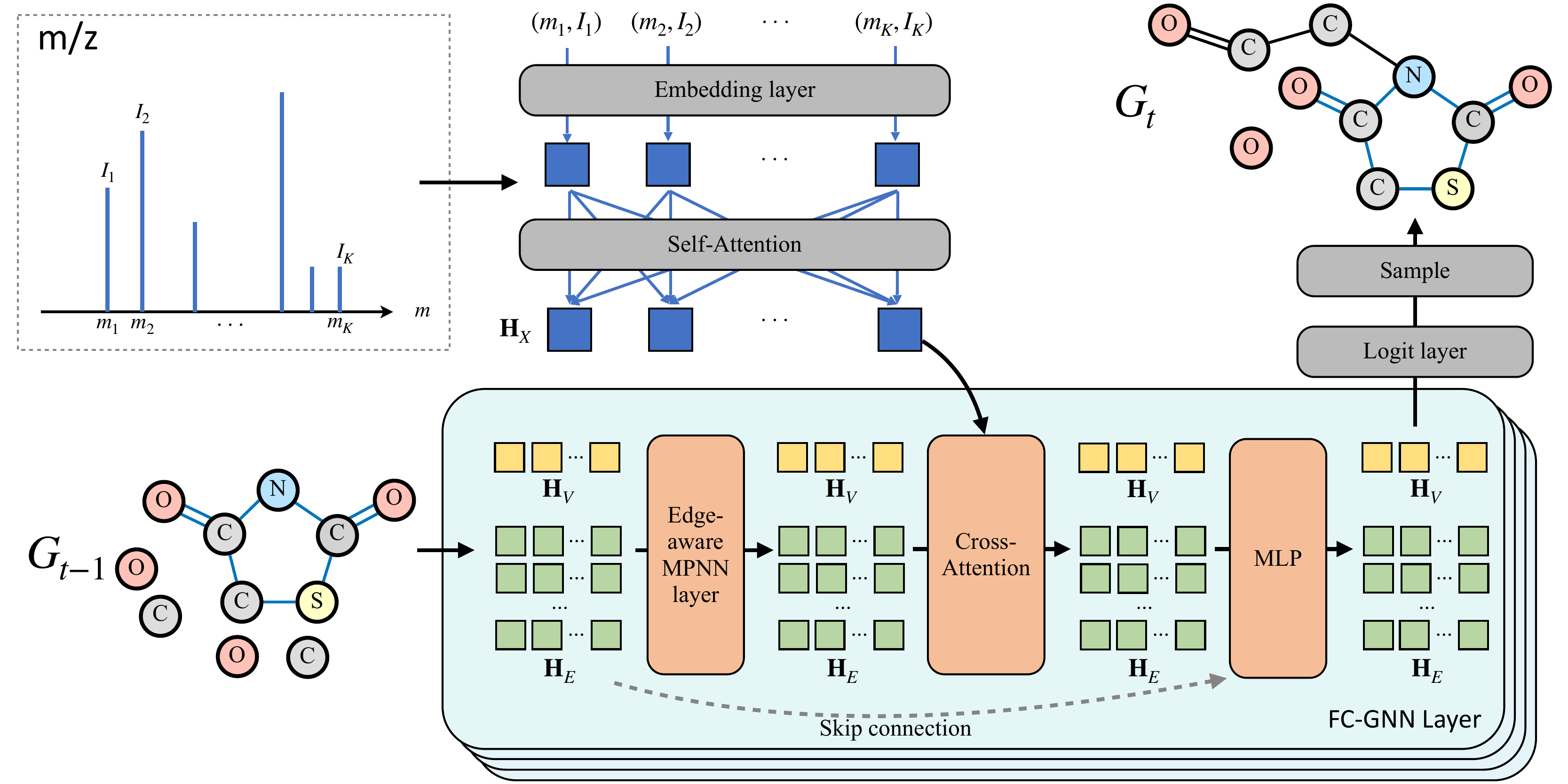}
    \caption{Overview of the MADGEN model framework. The input consists of m/z peaks and intensities \((m, I)\), which are passed through an MLP for embedding. These embeddings are processed through self-attention and combined with the molecular graph's node and edge embeddings via cross-attention. The node and edge embeddings are updated iteratively using an edge-aware message-passing neural network(MPNN) and fully-connected graph neural network (FC-GNN) layers. The final molecular structure is sampled after the last time step via a logit layer, aligning with the mass spectral data.
}
\label{fig:madgen}
\end{figure}

\subsubsection{Classifier-free Guidance from Mass Spectrum} We introduce the mass spectrum \( X^\mathrm{ms} \) as an additional conditioning term to refine the search space during the generation of \( G \) from \( S \). The neural network \( \mathrm{nn}_\theta(\cdot) \) is designed to condition on \( X \) when computing the logits.

To integrate spectrum information throughout the generation process, we utilize classifier-free guidance (CFG)~\citep{ho2022classifier}. At each inference step, for each edge entry, we compute the logit \( \ell_c \) conditioned on the spectrum \( X \), and the logit \( \ell_u \) without conditioning. The final logit \( \ell_g \) is then obtained by combining the two using a guidance scale \( \lambda_t \):
\begin{align}
    \ell_g = (1+\lambda_t)\ell_c - \lambda_t\ell_u.
\end{align}
During training, we randomly remove the spectrum condition with a probability of 0.1 to enable CFG. Since CFG tends to prioritize generation quality over diversity, increasing \( \lambda_t \) helps reduce the search space and improves the success rate of generating target molecules based on the given spectrum.

We provide further details on how the CFG techniques are integrated into our framework (see Figure~\ref{fig:madgen}), particularly within the network architecture \( \mathrm{nn}_\theta(\cdot) \). We treat the graph as fully connected, where non-edges are considered a specific type of edge, and apply a fully connected graph neural network (FC-GNN) to compute on this structure. The detailed design of the FC-GNN is provided in the Appendix~\ref{app:architectures}. Two key components to highlight are the encoding of the mass spectrum \( X \) and the conditioning mechanism.
\vspace{-0.8em}

\paragraph{Mass spectrum encoding as tokenization.} We represent the $X$ as a set of peaks $\{P_1,\ldots,P_K\}$, where $P_k=(M_k,I_k)$ is the m/z and intensity values. We encode each peak into a embedding vector  via an MLP. We then use a self-attention module to boost the information flow among the peak representations.  The full computation is as follow:
\begin{align}
    \mathbf{H}_X = \mathrm{Self}\text{-}\mathrm{Attention}(\mathbf{h}'_1,\ldots,\mathbf{h}'_K),~\mathbf{h}'_k=\mathrm{concat}(\mathrm{MLP}(M_k),~\mathrm{MLP}(I_k)).
\end{align}
This approach results in a variable-length representation of the mass spectrum, \(\mathbf{H}_X\), where each peak representation \(\mathbf{h}_k\) aligns with potential substructures in the molecule. By retaining these individual peak representations, the model is better able to guide the generation of subgraphs that correspond to molecular fragments consistent with the observed spectral data.

\vspace{-0.8em}

\paragraph{Spectrum conditioning via cross-attention.} We map $\mathbf{H}_X$ to each message passing layer of FC-GNN via cross-attention. Since there are intermediate representations for both nodes and edges, we explore three cross-attention paradigms: node-only attention, edge-only attention and both. We replace the $\mathbf{H}_X$ with a learnable embedding when the spectrum data are removed.

\section{Experiments}\label{sec4}


\begin{table}[t]
\centering
\small
\begin{tabular}{lccccccccc}
\toprule
\multirow{2}{*}{Dataset}\!\!\!&\!\!\!\multirow{2}{*}{\#Spec.}\!\!\!&\!\!\!\multirow{2}{*}{\#Mol.}\!\!\!&\!\!\!\multirow{2}{*}{\#Scaf.}\!\!\!&\!\!\!\multirow{2}{*}{\#Free atoms}& \multicolumn{3}{c}{per Mol.} & \multicolumn{2}{c}{per Scaf.}\\
\cmidrule(r){6-8} \cmidrule(r){9-10}&&&&&\#Node\!&\!\#Edge\!&\!\#Spec.\!&\!\#Node\!&\!\#Mol.\\\midrule
{NIST23} & 689,358 & 40,934 & 13,560 & 6.6 & 21.0 & 22.4  & 16.84 & 14.4 & 3.02\\
{CANOPUS} & 6,618 & 6,618 & 3,137 & 8.5 & 28.6 & 31.0 & 1.00 & 20.1 & 2.11 \\
{MassSpecGym} & 231,104 & 31,602 & 15,649 & 10.9 & 25.8 & 27.4 & 6.60 & 22.1 & 2.02\\
\bottomrule
\end{tabular}%

\caption{Statistics on the three datasets, including the number of molecules (Mol.), spectra (Spec.), scaffolds (Scaf.), average number of free atoms, and average statistics per molecule and per scaffold.}
\label{tab:datasets}
\end{table}


\subsection{Datasets}\label{subsec4.1}

We evaluate the performance of MADGEN on three datasets (Table \ref{tab:datasets}).  The NIST23  dataset \citep{nist23} is curated by the National Institute of Standards and Technology to provide reference spectral data for a wide range of chemical molecular standards to support research and development. It is available for purchase. Each molecule is measured using various mass spectrometry instruments, and under various instrument settings, thus contributing to the high number of spectra/molecule.  The CANOPUS dataset is the smallest dataset, and it was designed to train and evaluate the CANOPUS tool \citep{duhrkop2021systematic}, which predicts compound classes, e.g., alcohols, phenol ethers, and others, from spectra. It has a 1:1 spectra to molecule ratio. It was used recently to benchmark other metabolomics tools such as MIST\citep{goldman2023annotating} and ESP\citep{li2024ensemble}. The newly developed MassSpecGym benchmark dataset \citep{bushuiev2024mass} is collected  from  many public reference spectral databases and curated uniformly. The MassSpecGym is the largest publicly available labeled mass spectra dataset. For all three datasets, few molecules shared the same scaffold.

All datasets were preprocessed by normalizing the intensities of the MS/MS spectra and removing low-intensity peaks below a predefined threshold to reduce noise. The NIST23 and CANOPUS datasets were  split  into training, validation, and test sets based on the scaffold, ensuring that scaffolds are unique to each split. This split prevents data leakage and ensures robust evaluation of model performance. For MassSpecGym, we utilized the split suggested by the benchmark \citep{bushuiev2024mass}, which is based on the Maximum Common Edge Substructure (MCES).  This split allows assessing the model generalization on novel molecules.


\subsection{Experimental Setup and Evaluation Metrics}\label{subsec4.2}
The model was trained using a graph transformer with 5 layers and 50 diffusion steps. We employed the AdamW optimizer with a learning rate of \(1 \times 10^{-5}\). Full training details and hyperparameters can be found in Appendix~\ref{app:hyperparameters}.
For candidate pool selection, the following approaches were employed:
\begin{itemize}
    \item \textbf{NIST23 and CANOPUS}: all candidate molecules were retrieved from PubChem using the chemical formula as a query, ensuring comprehensive coverage of possible structures.
    \item \textbf{MassSpecGym}: the candidate pool consists of 256 molecules for each test molecule. These candidates are selected based on the molecular formula provided by the MassSpecGym dataset. The target molecule is removed from the candidate pool.
\end{itemize}
The performance of the model is evaluated using the following metrics endorsed for model evaluation for the MassSpecGym benchmark:
\vspace{-0.8em}
\begin{itemize}
    \item \textbf{Top-\textit{k} accuracy:} Measures the likelihood of the generating the true target structure among the top-\textit{k} generated molecules. We report the results for \textit{k=1,10}. The generated molecules are ranked by the probabilistic nature of the model.
    \vspace{-0.3em}

    \item \textbf{Tanimoto Similarity:} This metric evaluates the similarity between the generated  structures and the ground truth molecules using molecular fingerprints. Higher Tanimoto similarity indicates that the predicted structure closely resembles the correct structure. We extracted fingerprint representations based on the Morgan algorithm \citep{morgan1965generation} using the RDKit toolkit \citep{rdkit}. The Morgan fingerprints are computed for radius 2 and 2048 bits.
    \vspace{-0.3em}

    \item \textbf{Maximum Common Edge Substructure (MCES):} This metric is the edit distance between two molecules, and reflects the similarity of the largest common substructure between  generated and ground truth molecules\citep{Kretschmer2023.03.27.534311}.
    \vspace{-0.3em}

    \item \textbf{Scaffold Prediction Accuracy (SPA):} In  the scaffold prediction task, we assess how well the model predicts the core scaffold of the molecule compared to the ground truth scaffold.
\end{itemize}


\subsection{Results}

Our experiments, summarized in Table~\ref{tab:result}
, evaluate model performance on three datasets: NIST23, CANOPUS, and MassSpecGym, using both predictive and oracle retrievers. For the scaffold prediction task, we report a Scaffold Prediction Accuracy (SPA) for the predictive retriever ranging from 13.2\% to 40.3\%. Notably, the NIST23 dataset achieves the highest SPA of 40.3\%, reflecting its lower scaffold diversity compared to CANOPUS and MassSpecGym, which have more complex scaffolds.

The metrics for the scaffold-based generation task reveal that the low scaffold prediction accuracy of the predictive retriever constrains molecular generation performance. For instance, on the NIST23 dataset, the predictive retriever yields a top-1 accuracy of 4.6\%, while for CANOPUS and MassSpecGym, the top-1 accuracies is 2.10 \% and 1.31\%. Despite these challenges, the predictive retriever demonstrates moderate performance improvements compared to baseline methods like Spec2Mol and random generation.

In contrast, the oracle retriever, which has access to the correct scaffold, dramatically boosts performance. On NIST23, MADGEN achieves a top-1 accuracy of 49.0\% and a top-10 accuracy of 65.5\%, demonstrating the model's capacity to generate accurate molecular structures if the scaffold is known. Similarly, when using the oracle retriever, the performance on CANOPUS and MassSpecGym is significantly higher than the predictive retriever, with top-1 accuracies of 18.7\% and 10.5\%, respectively, showing the clear advantage of having access to correct scaffold information. Importantly, MADGEN outperforms the best published state-of-the-art performance (last row in Table~\ref{tab:result}) reported for the  MassSpecGym dataset \citep{bushuiev2024mass} when using  random chemical generation. The high top-1 and Top10 accuracy for the NIST23 dataset can be attributed to its smaller number of free atoms. MADGEN's task of completing the target molecule by adding edges to these free atoms is easier with a smaller number of free atoms. CANOPUS has the highest number of average free atoms, and the lowest top-1 and top-10 accuracies. 

Baseline methods like Spec2Mol and MSNovelist are also included in the comparison. As shown in Table~\ref{tab:result}, MSNovelist results are limited to accuracy metrics, as other measures are not available. The "-" in the table denotes this lack of data, while underlined values highlight the best results achieved by predictive retrievers, serving as a benchmark against the oracle retriever.

\begin{table}[!htbp]
\resizebox{\textwidth}{!}{
\begin{tabular}{lccccccc}
\toprule
\multirow{2}{*}{Retriever} & & \multicolumn{3}{c}{Top1} & \multicolumn{3}{c}{Top10} \\
\cmidrule(r){3-5} \cmidrule(r){6-8}
& SPA$\uparrow$ & Accuracy$\uparrow$ & Similarity$\uparrow$ & MCES$\downarrow$ & Accuracy$\uparrow$ & Similarity$\uparrow$ & MCES$\downarrow$ \\
\midrule
\multicolumn{8}{c}{NIST}
\\\midrule
Spec2Mol & - & 0.0\% & 0.10 & \underline{20.88} & 0.0\% & 0.12 & \underline{13.66} \\
MSNovelist & - & 0.0\% & - & - & 0.0\% & - & - \\
\rowcolor{lgray}
MADGEN$_\text{Pred.}$& 40.3\% &		\underline{4.6\%}          &   \underline{0.11}   &    72.38     &     \underline{7.3\%}     &    \underline{0.18}     &     69.34\\
\rowcolor{lgray}
MADGEN$_\text{Oracle}$ & 100\% & \textbf{49.0\%} & \textbf{0.63} & \textbf{18.48} & \textbf{65.5\%} & \textbf{0.80} & \textbf{3.88}\\
\midrule
\multicolumn{8}{c}{CANOPUS}
\\\midrule
Spec2Mol & - & 0.0\% & 0.09 & 38.97 & 0.0\% & 0.14 & 23.97\\
MSNovelist & - & 0.0\% & - & - & 0.0\% & - & - \\
\rowcolor{lgray}
MADGEN$_\text{Pred.}$ & 17.7\%	&	\underline{2.10\%}           &  \underline{0.22}     &  \underline{20.56}       &   \underline{2.39\%}     &    \underline{0.27}       &   \underline{12.69}
\\
\rowcolor{lgray}
MADGEN$_\text{Oracle}$ & 100\% & \textbf{18.7\%} & \textbf{0.51} & \textbf{9.72} & \textbf{22.2\%} & \textbf{0.59} & \textbf{4.44}\\
\midrule
\multicolumn{8}{c}{MassSpecGym}
\\\midrule
Rand. Gen. & - & 0.0\% & 0.08 & \underline{21.11} & 0.0\% & 0.11 & 18.25 \\
SMILES Transformer & - & 0.0\% & 0.03 & 79.39 & 0.0\% & 0.10 & 52.13 \\
SELFIES Transformer & - & 0.0\% & 0.08 & 38.88 & 0.0\% & 0.13 & 26.87 \\
Spec2Mol & - & 0.0\% & 0.09 & 45.89 & 0.0\% & 0.13 & 32.60 \\
MSNovelist & - & 0.0\% & - & - & 0.0\% & - & - \\
\rowcolor{lgray}
MADGEN$_\text{Pred.}$ & 13.2\% &	\underline{1.31\%}   &         \underline{0.20}    &   27.47       &   \underline{1.54\%}   &      \underline{0.26}     &    \underline{16.84}
\\
\rowcolor{lgray}
MADGEN$_\text{Oracle}$ & 100\% & \textbf{10.5\%} & \textbf{0.43} & \textbf{16.27} & \textbf{12.4\%} & \textbf{0.53} & \textbf{7.08}\\

\bottomrule
\end{tabular}}
\caption{Performance metrics for various datasets using both predictive and oracle retrievers. The table presents top-1 and top-10 accuracy, Tanimoto similarity, and Maximum Common Edge Substructure (MCES) scores. Best performance for each dataset is \textbf{bold}. The second-best performance for each dataset is \underline{underlined}}\label{tab:result}

\end{table}

\subsection{Ablation Study on Conditioning Mechanism}
We conducted an ablation study to assess the impact of different encoding strategies, conditioning methods, and the use of CFG (Table~\ref{tab:ablation_study}) on the performance of MADGEN. Conditioning using the  tokenization + cross-attention mechanism significantly improves the model performance. We believe this is because such mass spectra encoding is more efficient in encoding peak information without further compression. Further, through cross-attention, nodes and edges are able to query peaks of relevant importance. Importantly, upon introducing the self-attention into the mass spectrum encoder, a dramatic performance gain is observed. The self-attention significantly enhances mass spectra representations. We observed further performance gains using  CFG on node or node and edge (both). Node-only CFG yields the best performance among all settings.

\begin{table}[!htbp]
\centering
\begin{tabular}{cc|cccc}
\toprule
Encoding strategy & Conditioning strategy& Accuracy & Similarity & MCES \\
\midrule
Binning + MLP & Concatenation &  4.30\% & 0.068 & 87.89 \\

Tokenization & Cross-Attn& 13.0\% & 0.304 & 55.25\\

Tokenization + Self-Attn & Cross-Attn & 42.5\% & 0.642 & 23.90\\

Tokenization + Self-Attn & Cross-Attn + CFG (edge) & 42.0\% & 0.632 & 25.01\\

Tokenization + Self-Attn & Cross-Attn + CFG (node) & \textbf{49.0\%} & \textbf{0.694} & \textbf{18.48}\\

Tokenization + Self-Attn & Cross-Attn + CFG (both) & 45.9\% & 0.667 & 21.89\\
 \bottomrule
\end{tabular}
\caption{Ablation study results comparing different encoding strategies (Binning + MLP, Tokenization, Tokenization + Self-Attention) and conditioning strategies (Concatenation, Cross-Attention, Cross-Attention + CFG). The metrics evaluated are Accuracy (\%), Tanimoto Similarity, and Maximum Common Edge Substructure (MCES). The best results were obtained using Tokenization + Self-Attention with Cross-Attention + CFG (node).}

\label{tab:ablation_study}
\end{table}
\vspace{-1em}

\subsection{Sensitivity Analysis of Free Atom Numbers on Accuracy}
\begin{figure}[!htbp]
    \centering
    \small
    \begin{tabular}{cc}
             \includegraphics[width=0.45\linewidth]{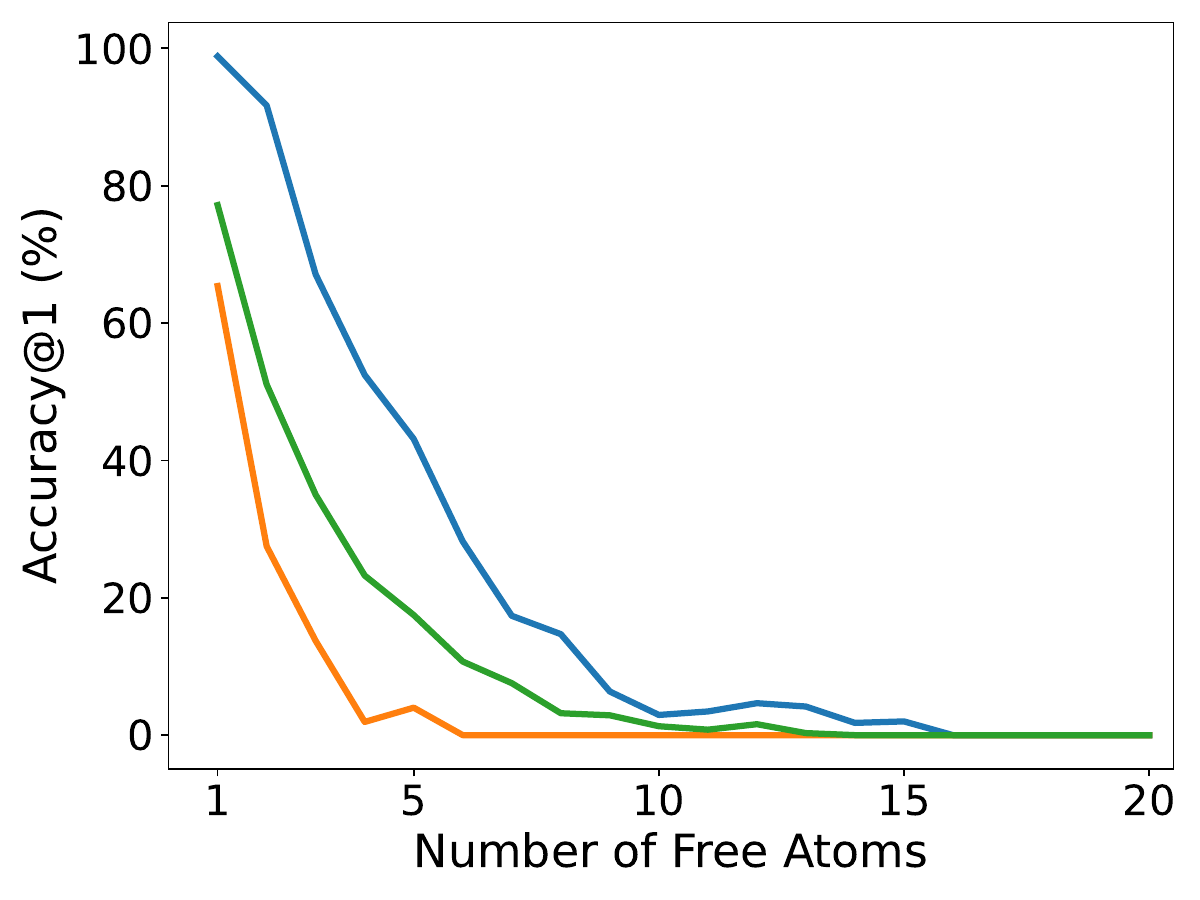} &
             \includegraphics[width=0.45\linewidth]{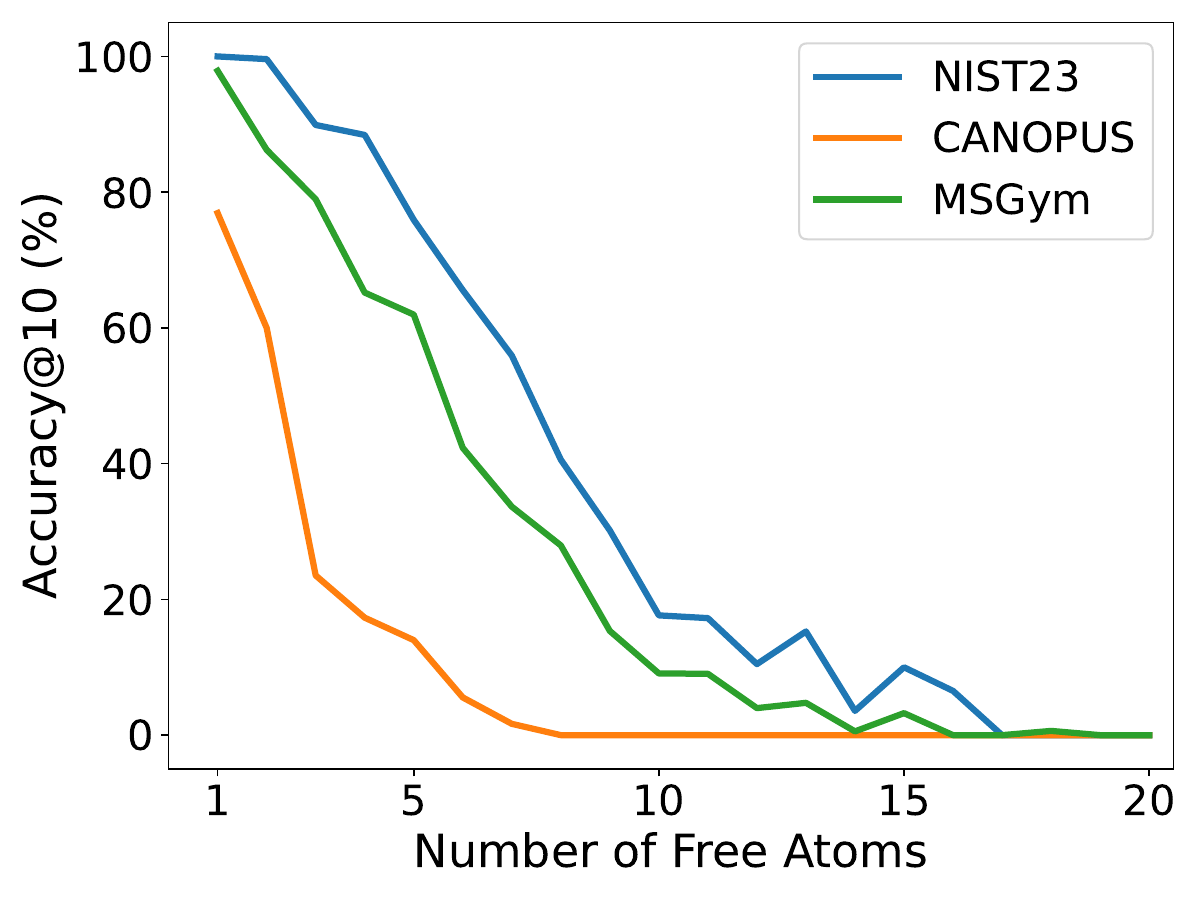}\\
             (a) Accuracy@1 vs Number of Free Atoms & (b) Accuracy@10 vs Number of Free Atoms 
    \end{tabular}
    \caption{Accuracy vs Number of Free Atoms: With more free atoms for MADGEN to connect to the scaffold, the complexity of the generative trajectory increases, leading to a worse predictive accuracy.}
    \label{fig:acc-vs-fa}
\end{figure}

We analyze how the number of free atoms will affects the generation accuracy of MADGEN. Figure~\ref{fig:acc-vs-fa} shows MADGEN's accuracy@1 and accuracy@10 on different number of free atoms across three datasets. We can observe that having more free atoms yields worse predictive accuracy, which is as expected as the learning complexity increases.

\vspace{-1em}
\section{Conclusion \& Future Work}
De novo annotation of mass spectrometry data is notoriously difficult, with a current best accuracy of 0\% on the MassSpecGym dataset. 
MADGEN offers a novel two-stage  framework for spectra-guided de novo annotation. The first stage, scaffold retrieval, is a new problem formulation whose solution can provide partial insight in regard to the molecular backbone of the measured spectra. Such insights may shed light on the molecule's class or properties. Our results show that this problem is challenging, achieving a scaffold prediction accuracy of 13.2\%-40.3\% for the three datasets. The second stage, de novo generation from an existing scaffold showed excellent results when using an oracle scaffold predictor, achieving an accuracy of 10.5\%-49\% across the three dataset. For the MassSpecGym benchmark, we achieved an accuracy of 2.10\% and 1.31\%. As with other tools, e.g., \citep{goldman2023annotating}, we conjecture that performance of MADGEN can be increased by incorporating additional data in the form of peak chemical formulae  or molecular properties that correlate with fragmentation patterns. Potentially, the scaffold problem can be made easier if larger more distinct scaffold structures were utilized instead of the Murcko  scaffold used herein. A bigger scaffold can in turn facilitate the de novo generation task. Further, an end-to-end MADGEN may reduce the compounding of errors across the two stages. 

\subsubsection*{Acknowledgments}
Research reported in this publication was supported by the National Institute of General Medical Sciences of the National Institutes of Health under award number R35GM148219. The content is solely the responsibility of the authors and does not necessarily represent the official views of the NIH. Chen and Liu are supported by the NSF CAREER Award 2239869.

\newpage
\bibliography{iclr2025_conference}

\begin{thebibliography}{41}
\providecommand{\natexlab}[1]{#1}
\providecommand{\url}[1]{\texttt{#1}}
\expandafter\ifx\csname urlstyle\endcsname\relax
  \providecommand{\doi}[1]{doi: #1}\else
  \providecommand{\doi}{doi: \begingroup \urlstyle{rm}\Url}\fi

\bibitem[Austin et~al.(2021)Austin, Johnson, Ho, Tarlow, and Van Den~Berg]{d3pm}
Jacob Austin, Daniel~D Johnson, Jonathan Ho, Daniel Tarlow, and Rianne Van Den~Berg.
\newblock Structured denoising diffusion models in discrete state-spaces.
\newblock \emph{Advances in Neural Information Processing Systems}, 34:\penalty0 17981--17993, 2021.

\bibitem[Bain et~al.(2022)Bain, Nagrani, Varol, and Zisserman]{bain2022clip}
Max Bain, Arsha Nagrani, G{\"u}l Varol, and Andrew Zisserman.
\newblock A clip-hitchhiker's guide to long video retrieval.
\newblock \emph{arXiv preprint arXiv:2205.08508}, 2022.

\bibitem[Bemis \& Murcko(1996)Bemis and Murcko]{bemis1996properties}
Guy~W Bemis and Mark~A Murcko.
\newblock The properties of known drugs. 1. molecular frameworks.
\newblock \emph{Journal of medicinal chemistry}, 39\penalty0 (15):\penalty0 2887--2893, 1996.

\bibitem[Bushuiev et~al.(2024)Bushuiev, Bushuiev, de~Jonge, Young, Kretschmer, Samusevich, Heirman, Wang, Zhang, Dührkop, Ludwig, Haupt, Kalia, Brungs, Schmid, Greiner, Wang, Wishart, Liu, Rousu, Bittremieux, Rost, Mak, Hassoun, Huber, van~der Hooft, Stravs, Böcker, Sivic, and Pluskal]{bushuiev2024mass}
Roman Bushuiev, Anton Bushuiev, Niek de~Jonge, Adamo Young, Fleming Kretschmer, Raman Samusevich, Janne Heirman, Fei Wang, Luke Zhang, Kai Dührkop, Marcus Ludwig, Nils Haupt, Apurva Kalia, Corinna Brungs, Robin Schmid, Russell Greiner, Bo~Wang, David Wishart, Liping Liu, Juho Rousu, Wout Bittremieux, Hannes Rost, Tytus Mak, Soha Hassoun, Florian Huber, Justin~J.J. van~der Hooft, Michael Stravs, Sebastian Böcker, Josef Sivic, and Tomáš Pluskal.
\newblock Massspecgym: A benchmark for the discovery and identification of molecules.
\newblock \emph{Advances in Neural Information Processing Systems}, 2024.

\bibitem[Butler et~al.(2023)Butler, Frandsen, Lightheart, Bargh, Taylor, Bollerman, Kerby, West, Voronov, Moon, et~al.]{butler2023ms2mol}
Thomas Butler, Abraham Frandsen, Rose Lightheart, Brian Bargh, James Taylor, TJ~Bollerman, Thomas Kerby, Kiana West, Gennady Voronov, Kevin Moon, et~al.
\newblock Ms2mol: A transformer model for illuminating dark chemical space from mass spectra.
\newblock \emph{ChemRxiv. 2023; doi:10.26434/chemrxiv-2023-vsmpx-v2}, 2023.

\bibitem[Chen et~al.(2023)Chen, He, Han, and Liu]{chen2023efficient}
Xiaohui Chen, Jiaxing He, Xu~Han, and Li-Ping Liu.
\newblock Efficient and degree-guided graph generation via discrete diffusion modeling.
\newblock \emph{arXiv preprint arXiv:2305.04111}, 2023.

\bibitem[Davis()]{MoNA}
UC~Davis.
\newblock {MassBank of North America}.
\newblock \url{https://mona.fiehnlab.ucdavis.edu/}.
\newblock URL \url{https://mona.fiehnlab.ucdavis.edu/}.

\bibitem[Duan et~al.(2024)Duan, Liu, Du, Chen, Zhao, Jia, Gomes, Theodorou, and Kulik]{duan2024react}
Chenru Duan, Guan-Horng Liu, Yuanqi Du, Tianrong Chen, Qiyuan Zhao, Haojun Jia, Carla~P Gomes, Evangelos~A Theodorou, and Heather~J Kulik.
\newblock React-ot: Optimal transport for generating transition state in chemical reactions.
\newblock \emph{arXiv preprint arXiv:2404.13430}, 2024.

\bibitem[D{\"u}hrkop et~al.(2015)D{\"u}hrkop, Shen, Meusel, Rousu, and B{\"o}cker]{duhrkop2015searching}
Kai D{\"u}hrkop, Huibin Shen, Marvin Meusel, Juho Rousu, and Sebastian B{\"o}cker.
\newblock Searching molecular structure databases with tandem mass spectra using csi: Fingerid.
\newblock \emph{Proceedings of the National Academy of Sciences}, 112\penalty0 (41):\penalty0 12580--12585, 2015.

\bibitem[D{\"u}hrkop et~al.(2019)D{\"u}hrkop, Fleischauer, Ludwig, Aksenov, Melnik, Meusel, Dorrestein, Rousu, and B{\"o}cker]{duhrkop2019sirius}
Kai D{\"u}hrkop, Markus Fleischauer, Marcus Ludwig, Alexander~A Aksenov, Alexey~V Melnik, Marvin Meusel, Pieter~C Dorrestein, Juho Rousu, and Sebastian B{\"o}cker.
\newblock Sirius 4: a rapid tool for turning tandem mass spectra into metabolite structure information.
\newblock \emph{Nature methods}, 16\penalty0 (4):\penalty0 299--302, 2019.

\bibitem[D{\"u}hrkop et~al.(2021)D{\"u}hrkop, Nothias, Fleischauer, Reher, Ludwig, Hoffmann, Petras, Gerwick, Rousu, Dorrestein, et~al.]{duhrkop2021systematic}
Kai D{\"u}hrkop, Louis-F{\'e}lix Nothias, Markus Fleischauer, Raphael Reher, Marcus Ludwig, Martin~A Hoffmann, Daniel Petras, William~H Gerwick, Juho Rousu, Pieter~C Dorrestein, et~al.
\newblock Systematic classification of unknown metabolites using high-resolution fragmentation mass spectra.
\newblock \emph{Nature biotechnology}, 39\penalty0 (4):\penalty0 462--471, 2021.

\bibitem[Fang et~al.(2021)Fang, Xiong, Xu, and Chen]{fang2021clip2video}
Han Fang, Pengfei Xiong, Luhui Xu, and Yu~Chen.
\newblock Clip2video: Mastering video-text retrieval via image clip.
\newblock \emph{arXiv preprint arXiv:2106.11097}, 2021.

\bibitem[Goldman et~al.(2023)Goldman, Wohlwend, Stra{\v{z}}ar, Haroush, Xavier, and Coley]{goldman2023annotating}
Samuel Goldman, Jeremy Wohlwend, Martin Stra{\v{z}}ar, Guy Haroush, Ramnik~J Xavier, and Connor~W Coley.
\newblock Annotating metabolite mass spectra with domain-inspired chemical formula transformers.
\newblock \emph{Nature Machine Intelligence}, 5\penalty0 (9):\penalty0 965--979, 2023.

\bibitem[Green et~al.(2021)Green, Koes, and Durrant]{green2021deepfrag}
Harrison Green, David~R Koes, and Jacob~D Durrant.
\newblock Deepfrag: a deep convolutional neural network for fragment-based lead optimization.
\newblock \emph{Chemical Science}, 12\penalty0 (23):\penalty0 8036--8047, 2021.

\bibitem[Hendriksen et~al.(2022)Hendriksen, Bleeker, Vakulenko, Van~Noord, Kuiper, and De~Rijke]{hendriksen2022extending}
Mariya Hendriksen, Maurits Bleeker, Svitlana Vakulenko, Nanne Van~Noord, Ernst Kuiper, and Maarten De~Rijke.
\newblock Extending clip for category-to-image retrieval in e-commerce.
\newblock In \emph{European Conference on Information Retrieval}, pp.\  289--303. Springer, 2022.

\bibitem[Ho \& Salimans(2022)Ho and Salimans]{ho2022classifier}
Jonathan Ho and Tim Salimans.
\newblock Classifier-free diffusion guidance.
\newblock \emph{arXiv preprint arXiv:2207.12598}, 2022.

\bibitem[Holdijk et~al.(2022)Holdijk, Du, Jaini, Hooft, Ensing, and Welling]{holdijk2022path}
Lars Holdijk, Yuanqi Du, Priyank Jaini, Ferry Hooft, Bernd Ensing, and Max Welling.
\newblock Path integral stochastic optimal control for sampling transition paths.
\newblock In \emph{ICML 2022 2nd AI for Science Workshop}, 2022.

\bibitem[Hu et~al.(2016)Hu, Stumpfe, and Bajorath]{hu2016computational}
Ye~Hu, Dagmar Stumpfe, and Jürgen Bajorath.
\newblock Computational exploration of molecular scaffolds in medicinal chemistry: Miniperspective.
\newblock \emph{Journal of medicinal chemistry}, 59\penalty0 (9):\penalty0 4062--4076, 2016.

\bibitem[Igashov et~al.(2023)Igashov, Schneuing, Segler, Bronstein, and Correia]{igashov2023retrobridge}
Ilia Igashov, Arne Schneuing, Marwin Segler, Michael Bronstein, and Bruno Correia.
\newblock Retrobridge: Modeling retrosynthesis with markov bridges.
\newblock \emph{arXiv preprint arXiv:2308.16212}, 2023.

\bibitem[Kalia et~al.(2024)Kalia, Krishnan, and Hassoun]{kalia2024jestr}
Apurva Kalia, Dilip Krishnan, and Soha Hassoun.
\newblock Jestr: Joint embedding space technique for ranking candidate molecules for the annotation of untargeted metabolomics data.
\newblock \emph{arXiv preprint arXiv:2411.14464}, 2024.

\bibitem[Kanehisa et~al.(2021)Kanehisa, Furumichi, Sato, Ishiguro-Watanabe, and Tanabe]{kanehisa2021kegg}
Minoru Kanehisa, Miho Furumichi, Yoko Sato, Mari Ishiguro-Watanabe, and Mao Tanabe.
\newblock Kegg: integrating viruses and cellular organisms.
\newblock \emph{Nucleic acids research}, 49\penalty0 (D1):\penalty0 D545--D551, 2021.

\bibitem[Kim et~al.(2016)Kim, Thiessen, Bolton, Chen, Fu, Gindulyte, Han, He, He, Shoemaker, et~al.]{kim2016pubchem}
Sunghwan Kim, Paul~A Thiessen, Evan~E Bolton, Jie Chen, Gang Fu, Asta Gindulyte, Lianyi Han, Jane He, Siqian He, Benjamin~A Shoemaker, et~al.
\newblock Pubchem substance and compound databases.
\newblock \emph{Nucleic acids research}, 44\penalty0 (D1):\penalty0 D1202--D1213, 2016.

\bibitem[Kind et~al.(2018)Kind, Tsugawa, Cajka, Ma, Lai, Mehta, Wohlgemuth, Barupal, Showalter, Arita, et~al.]{kind2018identification}
Tobias Kind, Hiroshi Tsugawa, Tomas Cajka, Yan Ma, Zijuan Lai, Sajjan~S Mehta, Gert Wohlgemuth, Dinesh~Kumar Barupal, Megan~R Showalter, Masanori Arita, et~al.
\newblock Identification of small molecules using accurate mass ms/ms search.
\newblock \emph{Mass spectrometry reviews}, 37\penalty0 (4):\penalty0 513--532, 2018.

\bibitem[Kretschmer et~al.(2023)Kretschmer, Seipp, Ludwig, Klau, and B{\"o}cker]{Kretschmer2023.03.27.534311}
Fleming Kretschmer, Jan Seipp, Marcus Ludwig, Gunnar~W. Klau, and Sebastian B{\"o}cker.
\newblock Small molecule machine learning: All models are wrong, some may not even be useful.
\newblock \emph{bioRxiv}, 2023.
\newblock \doi{10.1101/2023.03.27.534311}.
\newblock URL \url{https://www.biorxiv.org/content/early/2023/03/27/2023.03.27.534311}.

\bibitem[Lei et~al.(2021)Lei, Li, Zhou, Gan, Berg, Bansal, and Liu]{lei2021less}
Jie Lei, Linjie Li, Luowei Zhou, Zhe Gan, Tamara~L Berg, Mohit Bansal, and Jingjing Liu.
\newblock Less is more: Clipbert for video-and-language learning via sparse sampling.
\newblock In \emph{Proceedings of the IEEE/CVF conference on computer vision and pattern recognition}, pp.\  7331--7341, 2021.

\bibitem[Li et~al.(2024)Li, Zhou~Chen, Kalia, Zhu, Liu, and Hassoun]{li2024ensemble}
Xinmeng Li, Yan Zhou~Chen, Apurva Kalia, Hao Zhu, Li-ping Liu, and Soha Hassoun.
\newblock An ensemble spectral prediction (esp) model for metabolite annotation.
\newblock \emph{Bioinformatics}, 40\penalty0 (8):\penalty0 btae490, 2024.

\bibitem[Li et~al.(2019)Li, Hu, Wang, Zhou, Zhang, and Liu]{li2019deepscaffold}
Yibo Li, Jianxing Hu, Yanxing Wang, Jielong Zhou, Liangren Zhang, and Zhenming Liu.
\newblock Deepscaffold: a comprehensive tool for scaffold-based de novo drug discovery using deep learning.
\newblock \emph{Journal of chemical information and modeling}, 60\penalty0 (1):\penalty0 77--91, 2019.

\bibitem[Litsa et~al.(2023)Litsa, Chenthamarakshan, Das, and Kavraki]{litsa2023end}
Eleni~E Litsa, Vijil Chenthamarakshan, Payel Das, and Lydia~E Kavraki.
\newblock An end-to-end deep learning framework for translating mass spectra to de-novo molecules.
\newblock \emph{Communications Chemistry}, 6\penalty0 (1):\penalty0 132, 2023.

\bibitem[Luo et~al.(2021)Luo, Ji, Zhong, Chen, Lei, Duan, and Li]{luo2021clip4clip}
Huaishao Luo, Lei Ji, Ming Zhong, Yang Chen, Wen Lei, Nan Duan, and Tianrui Li.
\newblock Clip4clip: An empirical study of clip for end to end video clip retrieval.
\newblock \emph{arXiv preprint arXiv:2104.08860}, 2021.

\bibitem[Ma et~al.(2022)Ma, Xu, Sun, Yan, Zhang, and Ji]{ma2022x}
Yiwei Ma, Guohai Xu, Xiaoshuai Sun, Ming Yan, Ji~Zhang, and Rongrong Ji.
\newblock X-clip: End-to-end multi-grained contrastive learning for video-text retrieval.
\newblock In \emph{Proceedings of the 30th ACM International Conference on Multimedia}, pp.\  638--647, 2022.

\bibitem[Morgan(1965)]{morgan1965generation}
Harry~L Morgan.
\newblock The generation of a unique machine description for chemical structures-a technique developed at chemical abstracts service.
\newblock \emph{Journal of chemical documentation}, 5\penalty0 (2):\penalty0 107--113, 1965.

\bibitem[{National Institute of Standards and Technology (NIST)}(2023)]{nist23}
{National Institute of Standards and Technology (NIST)}.
\newblock Nist 23 updates to the nist tandem and electron ionization spectral libraries, 2023.
\newblock URL \url{https://www.nist.gov/programs-projects/nist23-updates-nist-tandem-and-electron-ionization-spectral-libraries}.
\newblock Accessed: 2024-09-23.

\bibitem[Podda et~al.(2020)Podda, Bacciu, and Micheli]{podda2020deep}
Marco Podda, Davide Bacciu, and Alessio Micheli.
\newblock A deep generative model for fragment-based molecule generation.
\newblock In \emph{International conference on artificial intelligence and statistics}, pp.\  2240--2250. PMLR, 2020.

\bibitem[Radford et~al.(2021)Radford, Kim, Hallacy, Ramesh, Goh, Agarwal, Sastry, Askell, Mishkin, Clark, et~al.]{radford2021learning}
Alec Radford, Jong~Wook Kim, Chris Hallacy, Aditya Ramesh, Gabriel Goh, Sandhini Agarwal, Girish Sastry, Amanda Askell, Pamela Mishkin, Jack Clark, et~al.
\newblock Learning transferable visual models from natural language supervision.
\newblock In \emph{International conference on machine learning}, pp.\  8748--8763. PMLR, 2021.

\bibitem[RDKit, online()]{rdkit}
RDKit, online.
\newblock {RDK}it: Open-source cheminformatics.
\newblock \url{http://www.rdkit.org}.
\newblock [Online; accessed 11-April-2013].

\bibitem[Shrivastava et~al.(2021)Shrivastava, Swainston, Samanta, Roberts, Wright~Muelas, and Kell]{shrivastava2021massgenie}
Aditya~Divyakant Shrivastava, Neil Swainston, Soumitra Samanta, Ivayla Roberts, Marina Wright~Muelas, and Douglas~B Kell.
\newblock Massgenie: A transformer-based deep learning method for identifying small molecules from their mass spectra.
\newblock \emph{Biomolecules}, 11\penalty0 (12):\penalty0 1793, 2021.

\bibitem[Stravs et~al.(2022)Stravs, D{\"u}hrkop, B{\"o}cker, and Zamboni]{stravs2022msnovelist}
Michael~A Stravs, Kai D{\"u}hrkop, Sebastian B{\"o}cker, and Nicola Zamboni.
\newblock Msnovelist: de novo structure generation from mass spectra.
\newblock \emph{Nature Methods}, 19\penalty0 (7):\penalty0 865--870, 2022.

\bibitem[Wang et~al.(2016)Wang, Carver, Phelan, Sanchez, Garg, Peng, Nguyen, Watrous, Kapono, Luzzatto-Knaan, et~al.]{wang2016sharing}
Mingxun Wang, Jeremy~J Carver, Vanessa~V Phelan, Laura~M Sanchez, Neha Garg, Yao Peng, Don~Duy Nguyen, Jeramie Watrous, Clifford~A Kapono, Tal Luzzatto-Knaan, et~al.
\newblock Sharing and community curation of mass spectrometry data with global natural products social molecular networking.
\newblock \emph{Nature biotechnology}, 34\penalty0 (8):\penalty0 828--837, 2016.

\bibitem[Wei et~al.(2019)Wei, Belanger, Adams, and Sculley]{wei2019rapid}
Jennifer~N Wei, David Belanger, Ryan~P Adams, and D~Sculley.
\newblock Rapid prediction of electron--ionization mass spectrometry using neural networks.
\newblock \emph{ACS central science}, 5\penalty0 (4):\penalty0 700--708, 2019.

\bibitem[Zhu et~al.(2020)Zhu, Liu, and Hassoun]{zhu2020using}
H~Zhu, L~Liu, and S~Hassoun.
\newblock Using graph neural networks for mass spectrometry prediction. arxiv.
\newblock \emph{arXiv preprint arXiv:2010.04661}, 2020.

\bibitem[Zhu et~al.(2022)Zhu, Du, Wang, Xu, Zhang, Liu, and Wu]{zhu2022survey}
Yanqiao Zhu, Yuanqi Du, Yinkai Wang, Yichen Xu, Jieyu Zhang, Qiang Liu, and Shu Wu.
\newblock A survey on deep graph generation: Methods and applications.
\newblock In \emph{Learning on Graphs Conference}, pp.\  47--1. PMLR, 2022.

\end{thebibliography}
\bibliographystyle{iclr2025_conference}
\newpage
\appendix
\section{Appendix}

\subsection{Model Architectures and Algorithms}\label{app:architectures}

In this section, we describe the architecture of our proposed scaffold-conditioned molecular generation guided by mass spectra data. The process consists of two main stages: scaffold retrieval (Stage 1) and scaffold-conditioned molecular generation (Stage 2). The model integrates node, edge, and spectral features, updated iteratively through cross-attention and self-attention mechanisms.

\subsubsection{Stage 1: Scaffold Retrieval}
The first stage of the process involves predicting a scaffold that is most consistent with the input MS/MS spectrum. This stage utilizes a contrastive learning framework that aligns molecular graphs with their corresponding mass spectra. We use the `MLP\_BIN` model for encoding the spectral data.

\paragraph{Molecular encoder.}
We employ a Graph Neural Network (GNN) to encode the molecular structures:
\begin{itemize}
    \item \textbf{Node Features:} The molecular graph nodes (atoms) are encoded using GNN layers, where each node is associated with a feature vector that encodes atom type and other properties.
    \item \textbf{Edge Features:} Bonds between atoms are represented by edge features, which are also encoded by the GNN.
    \item \textbf{Graph Pooling:} The output node embeddings from the GNN are pooled using a MaxPooling layer to create a graph-level representation.
\end{itemize}

\paragraph{Spectral encoder (MLP\_BIN).}
For encoding the MS/MS spectra, we use the `MLP\_BIN` encoder:
\begin{itemize}
    \item The input spectra are represented as bins of mass-to-charge (m/z) ratios and intensities.
    \item The `MLP\_BIN` model processes these binned inputs through multiple fully connected layers, where each layer applies a ReLU activation and dropout to prevent overfitting.
    \item The output of the `MLP\_BIN` encoder is a vector representing the spectral data in an embedding space suitable for contrastive learning.
\end{itemize}

\paragraph{Interaction model.}
Once the molecular and spectral embeddings are computed, they are concatenated and passed through an interaction MLP, which predicts the interaction score between the molecule and the scaffold. The interaction score is used to rank candidate scaffolds. The molecular encoder and spectral encoder are trained jointly in a contrastive learning framework, where the goal is to align the embeddings of correct molecular-scaffold pairs.

\subsubsection{Stage 2: Scaffold-Conditioned Molecular Generation}
In Stage 2, the retrieved scaffold is used as the foundation for generating the full molecular structure, guided by the mass spectra data. This stage employs a \textbf{Graph Transformer} to integrate node, edge, and spectral features iteratively across multiple layers.

\paragraph{Input representation.}
The inputs to the Graph Transformer in this stage consist of:
\begin{itemize}
    \item \textbf{Node Features ($V$):} Each node represents an atom in the molecular scaffold, and the feature vector encodes atom type and properties.
    \item \textbf{Edge Features ($E$):} Bonds between atoms in the scaffold are represented as edge features.
    \item \textbf{Spectral Features ($S$):} The MS/MS spectra provide pairs of mass-to-charge (m/z) ratios and intensities.
\end{itemize}

\paragraph{Multi-head attention.}
Each layer of the Graph Transformer applies a \textbf{Node-Edge Block}, where both node and edge features are updated using attention mechanisms.
\begin{itemize}
    \item \textbf{Self-Attention:} The model computes queries, keys, and values for each node and edge, allowing it to focus on relevant parts of the molecular graph during the update process.
    \item \textbf{Cross-Attention:} Cross-attention between the node/edge features and the spectral features enables the generation process to be conditioned on the spectral data, ensuring that the generated molecular structure aligns with the spectra.
\end{itemize}

\paragraph{Feedforward networks.}
After the attention layers, a \textbf{FeedForward Network} processes the updated node and edge features, further refining the representations.

\paragraph{Layer normalization and residual connections.}
Each attention block is followed by \textbf{Layer Normalization} and residual connections to stabilize training and maintain information flow across the layers.

\paragraph{Final output.}
After the final transformer layer, the updated node and edge features are passed through an output MLP to generate the final molecular structure. This process ensures that the generated molecule is consistent with both the scaffold and the spectral data.

\subsection{Training Hyperparameters}\label{app:hyperparameters}

The model is trained with a batch size of 64 and employed 47 workers for data loading. The learning rate is set to \(2 \times 10^{-4}\), while weight decay is configured at \(1 \times 10^{-12}\). Training proceeds  for 2000 epochs, with the model logging progress every 40 steps.

A Markov bridge process with 100 steps is employed during training, and a cosine noise schedule is employed.

The model consists of 5 layers, with node, edge, and spectral features set at 64 dimensions each. The MLP hidden dimensions are configured to 256 for nodes, 128 for edges, and 256 for spectral features. The model also employs 8 attention heads for cross-attention and self-attention mechanisms. The feedforward dimensions are set to 256 for nodes, 128 for edges, and 128 for global features. This architecture enables efficient handling of both molecular structure and spectral data during training.

\subsection{Variation Distribution and ELBO}\label{app:derivation}
We first show the full derivation of the ELBO, which introduces a forward transition distribution $p(e_{t+1}|e_t,e_T)$ as the variational distribution. Then we discuss the formulation of the variational distribution. The derivation of the ELBO is as follow:
\begin{align}
\log p_\theta(G|S) &= \log \sum_{e_1:e_{T-1}}\prod_{t=0}^{T-1}p_\theta\big(e_{t+1}\big|e_t,\mathcal{E}^S,\mathcal{V}^G\big)\\
&=\log \sum_{e_1:e_{T-1}}\frac{p(e_{1:T-1}|e_0,e_T)}{p(e_{1:T-1}|e_0,e_T)}\prod_{t=0}^{T-1}p_\theta\big(e_{t+1}\big|e_t,\mathcal{E}^S,\mathcal{V}^G\big)\\
&\geq\mathbb{E}_{p(e_1:e_{T-1}|e_0,e_T)}\bigg[\log\frac{\prod_{t=0}^Tp_\theta\big(e_{t+1}\big|e_t,\mathcal{E}^S,\mathcal{V}^G\big)}{p(e_{0:T-1}|e_T)}\bigg]\\
&=\mathbb{E}_{p(e_0:e_{T-1}|e_0, e_T)}\bigg[\sum_{t=0}^T\log\frac{p_\theta\big(e_{t+1}\big|e_t,\mathcal{E}^S,\mathcal{V}^G\big)}{p(e_{t+1}|e_t,e_T)}\bigg]\\
&=\sum_{t=0}^{T-1}\mathbb{E}_{p(e_t,e_{t+1}|e_0,e_T)}\bigg[\log\frac{p_\theta\big(e_{t+1}\big|e_t,\mathcal{E}^S,\mathcal{V}^G\big)}{p(e_{t+1}|e_t,e_T)}\bigg]\\
&=\sum_{t=0}^{T-1}\mathbb{E}_{p(e_t|e_0,e_T)}\bigg[\mathbb{E}_{p(e_{t+1}|e_t,e_T)}\log\frac{p_\theta\big(e_{t+1}\big|e_t,\mathcal{E}^S,\mathcal{V}^G\big)}{p(e_{t+1}|e_t,e_T)}\bigg]\\
&=\sum_{t=0}^{T-1}\mathbb{E}_{p(e_t|e_0,e_T)}\bigg[-\mathrm{KL}\Big(p\big(e_{t+1}\big|e_{t},e_T\big)\big\|p_\theta\big(e_{t+1}\big|e_{t},\mathcal{E}^{S},\mathcal{V}^G\big)\Big)\bigg]\\
&=-T\mathbb{E}_{\mathcal{U}(t;0,T-1)}\mathbb{E}_{p(e_t|e_0,e_T)}\bigg[\mathrm{KL}\Big(p\big(e_{t+1}\big|e_{t},e_T\big)\big\|p_\theta\big(e_{t+1}\big|e_{t},\mathcal{E}^{S},\mathcal{V}^G\big)\Big)\bigg]\\
&:= \mathcal{L}_\theta(S,G)
\end{align}
The forward distribution defines a distribution of trajectories $e_{1:T-1}$ between $e_0$ and $e_T$. Note that $e_0$ is always 0 (non-edge) and is independent from $e_T$, so we have
\begin{align}
    p(e_{0:T-1}|e_T) = p(e_{1:T-1}|e_0,e_T).
\end{align}
This also satisfies the Markov property
\begin{align}
    p(e_{0:T-1}|e_T)=\prod_{t=0}^{T-1}p(e_{t+1}|e_t,e_T)=\prod_{t=0}^{T-1}\mathrm{Categorical}(e_{t+1};\mathbf{Q}_t(e_T)e_t).
\end{align}
The transition matrices $\mathbf{Q}_0,\ldots,\mathbf{Q}_{T-1}$ are $D\times D$ matrices, where
\begin{align}
    \mathbf{Q}_t(e_T)=\alpha_t\mathbf{I}_D + (1-\alpha_t)e_T\mathbf{I}_D^T.
\end{align}
$\alpha_0,\ldots,\alpha_{T-1}$ are scheduling parameters similar to~\citet{d3pm}. And $\mathbf{I}_D$ is a $D\times D$ identity matrix. With the defined transition matrices, the one-step transition probability $p(e_t|e_0,e_T)$ also has the closed form 
\begin{align}
    p(e_t|e_0,e_T)=\mathrm{Categorical}(e_{t+1};\bar{\mathbf{Q}}_t(e_T)e_t),~\bar{\mathbf{Q}}_t(e_T)=\prod_{\tau=0}^t\mathbf{Q}_\tau(e_T)
\end{align}
Now that both $p(e_t|e_0,e_T)$ and $p(e_{t+1}|e_t,e_T)$ can be derived in closed-form, we can directly optimize the ELBO $\mathcal{L}_\theta(S,G)$.

\subsection{Overall Workflow}
The two stages work together to form a scaffold-conditioned molecular generation system. In the first stage, the model retrieves a scaffold using contrastive learning and the `MLP\_BIN` spectral encoder, and in the second stage, the Graph Transformer uses this scaffold to generate a complete molecule, conditioned on the spectral data. This two-step approach ensures that the molecular generation process is both accurate and guided by experimentally observed spectra.
\end{document}